\documentclass{egpubl}
\usepackage{pg2026s}

\WsConferencePaper

\usepackage[T1]{fontenc}
\usepackage{dfadobe}

\usepackage{cite}
\usepackage{booktabs}
\usepackage{amsmath}
\usepackage{algorithm}
\usepackage{algpseudocode}
\usepackage[table]{xcolor}

\algnewcommand{\algorithmicto}{\textbf{to}}

\newif\ifshowrevisionchanges
\showrevisionchangesfalse
\definecolor{revisionred}{RGB}{190,0,0}
\definecolor{polishblue}{RGB}{0,70,180}
\DeclareRobustCommand{\revcolor}{\ifshowrevisionchanges\color{revisionred}\fi}
\DeclareRobustCommand{\rev}[1]{{\revcolor#1}}
\DeclareRobustCommand{\polish}[1]{{\ifshowrevisionchanges\color{polishblue}\fi#1}}

\BibtexOrBiblatex

\electronicVersion
\PrintedOrElectronic

\ifpdf
  \usepackage[pdftex]{graphicx}
\else
  \usepackage[dvips]{graphicx}
\fi

\usepackage{egweblnk}

\newif\ifshowpolishchanges
\showpolishchangesfalse
\newif\ifshowhistoryadditions
\showhistoryadditionsfalse
\definecolor{historypurple}{RGB}{128,0,128}
\DeclareRobustCommand{\polish}[1]{{\ifshowrevisionchanges\ifshowpolishchanges\color{polishblue}\fi\fi#1}}
\DeclareRobustCommand{\historyadd}[1]{{\ifshowrevisionchanges\ifshowhistoryadditions\color{historypurple}\fi\fi#1}}

\title[DReSG]%
      {DReSG: Diffusion Residuals for Stylized Gaussian Splatting}

\author[Z. Liu, W. Liu \& Y. Li]
{\parbox{\textwidth}{\centering Zhongliang Liu$^1$\orcid{0009-0005-0013-6201}, Wenjie Liu$^2$\orcid{0009-0008-0088-7701}, and Yang Li\rev{$^{2,3}$}\orcid{0000-0001-9427-7665}\thanks{Corresponding author}}
\\
{\parbox{\textwidth}{\centering
$^1$School of Software Engineering, East China Normal University, Shanghai, China\\
$^2$School of Computer Science and Technology, East China Normal University, Shanghai, China\\
\rev{$^3$School of Intelligent Interaction, East China Normal University, Shanghai, China}
}}
}

\begin{document}

\teaser{
  \includegraphics[width=\textwidth]{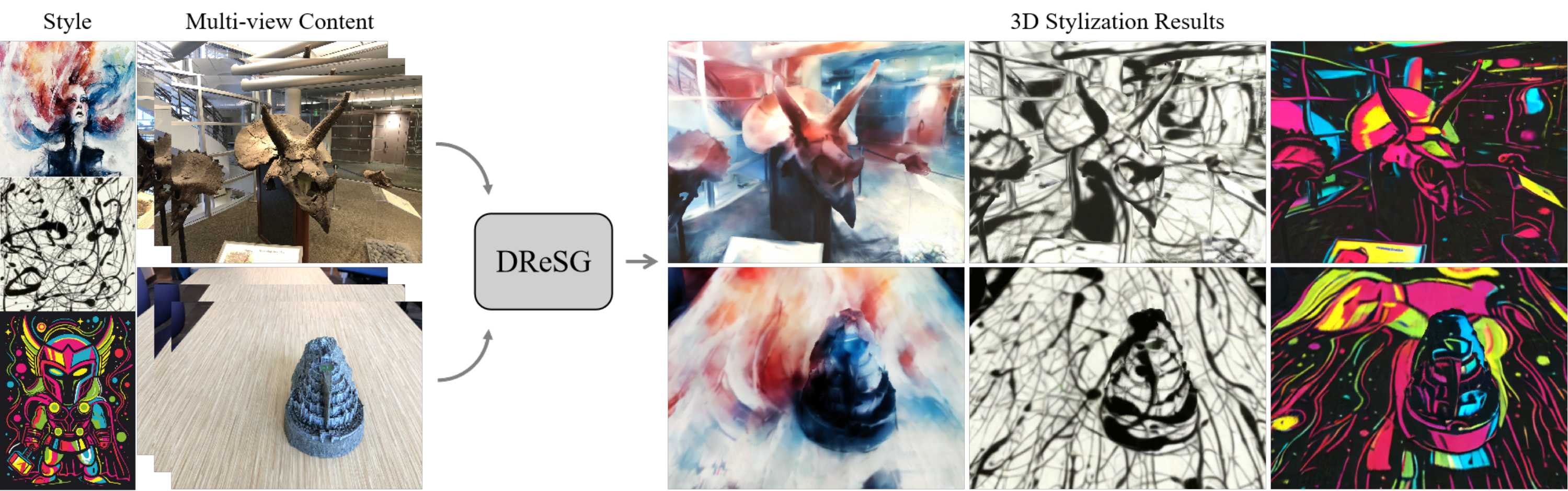}
  \centering
  \caption{DReSG fits diffusion-proposed stylization residuals into a reconstructed Gaussian scene, producing a renderable stylized Gaussian scene whose reference-specific appearance persists across novel views.}
\label{fig:teaser}
}

\maketitle
\begin{abstract}
Reference-guided stylization of scenes represented by 3D Gaussian Splatting
(3DGS) is important for efficient and controllable 3D content creation.
Existing VGG-feature-based 3D stylization methods provide stable rendered-view
optimization, but often
under-represent expressive reference style cues; diffusion models offer stronger
image priors, yet direct per-view or score-based diffusion guidance can lead to
view drift, local artifacts, and hard-to-control appearance updates. We present
\textbf{DReSG}, a 3D-grounded residual-feedback framework for stylized Gaussian
splatting. DReSG represents attention-guided diffusion proposals as residual
targets relative to the current render, and progressively absorbs these residuals
into a shared Gaussian scene through multi-view Gaussian feedback. To make this
feedback stable and controllable, DReSG modulates residual strength during target
construction and combines coverage-aware view selection with conflict-filtered
color updates during multi-view fitting. Extensive experiments demonstrate that
DReSG achieves competitive reference-guided stylization while better preserving
scene structure and cross-view stability.
\rev{Our project page is available at
\url{https://vpx-ecnu.github.io/DReSG-website/}.}

\ccsdesc[500]{Computing methodologies~Computer graphics}
\ccsdesc[500]{Computing methodologies~Rendering}
\ccsdesc[300]{Computing methodologies~Image processing}

\printccsdesc
\begin{keywords}
\rev{3D Gaussian Splatting; reference-guided stylization; diffusion models; multi-view consistency}
\end{keywords}
\end{abstract}  
\section{Introduction}
\label{sec:introduction}

3D style transfer aims to transfer the artistic appearance of a reference image
onto a 3D scene while preserving the scene's original structure and semantic
content~\cite{Zhang2022ARF,Zhang2024StylizedGS}. With the growing demand for
editable 3D content in virtual reality, film production, games, and digital
asset creation, 3D stylization has become an important tool for efficient and
controllable scene authoring. Compared with 2D image stylization, a
stylized Gaussian scene must transfer the reference appearance, preserve the
reconstructed structure, and \polish{maintain the 3D anchoring of local appearance
changes under camera motion}.

Most existing 3D stylization methods follow a rendered-view optimization
paradigm: they render images from a 3D representation and optimize the scene with
2D style supervision. \polish{This paradigm supports stable optimization and
multi-view constraints and has been widely used for both NeRF and Gaussian
stylization~\cite{Nguyen2022SNeRF,Zhang2022ARF,Liu2023StyleRF,
Liu2024StyleGaussian,Zhang2024StylizedGS}.} Many methods rely on
VGG-feature-based objectives, including Gram-based style losses, perceptual
feature reconstruction, adaptive normalization, or feature transforms
~\cite{Gatys2016Style,Johnson2016Perceptual,Huang2017AdaIN,Li2017WCT}. These
objectives effectively transfer global colors and local texture responses and
\polish{provide gradients that can be readily propagated through the
differentiable renderer}. However, VGG feature
objectives often express the reference style through local feature responses,
normalization parameters, or aggregate correlations. As a result, they can
under-represent spatially organized reference cues such as continuous strokes,
coherent contours, directional structures, and region-wise color treatment,
leading to \polish{weak stylization effects} or fragmented textures instead of a persistent
style update stored in the 3D scene.

More expressive image priors provide new opportunities for reference-guided 3D
stylization. CLIP-based guidance introduces high-level semantic or multimodal
image similarity~\cite{Radford2021CLIP,Howil2025CLIPGaussian}, but it is usually
used as a global matching objective rather than a local render-space target that
can be directly fitted by 3D rendering. Diffusion models are generative image
priors trained to synthesize images through iterative denoising
~\cite{Ho2020DDPM,Rombach2022LDM}. They can produce concrete stylization
proposals with richer semantic, structural, and local appearance changes than
VGG feature losses.

Despite this stronger prior, diffusion signals are not immediately stable
targets for 3DGS stylization. Per-view diffusion stylization can introduce
inconsistent colors, textures, or local structures across views
~\cite{Haque2023InstructNeRF2NeRF,Wang2024GaussianEditor,Chen2024DGE,
Yang2025FantasyStyle}, while directly
optimizing 3D parameters with diffusion scores or timestep-dependent gradients
~\cite{Poole2023DreamFusion,Wang2023SJC,Wang2023ProlificDreamer}
lacks an explicit render-space target and can produce saturated colors, local
artifacts, or \polish{poorly controlled appearance updates}. Our insight is to use
diffusion as a source of observable reference-aware stylization proposals, and
to convert these proposals into residual targets relative to the current render.
The key novelty is this render-relative residual-feedback formulation:
diffusion proposes reference-aware appearance changes, while the shared Gaussian
scene repeatedly absorbs the residuals that can be explained across views.

We propose \textbf{DReSG}, a 3D-grounded residual-feedback framework for
stylized Gaussian splatting. DReSG renders the current Gaussian scene, generates
attention-guided diffusion proposals, computes render-relative residual targets,
and fits these targets back to the shared Gaussian scene through differentiable
multi-view Gaussian rendering. To control the feedback strength, DReSG constructs
targets in RGB-logit space and applies SNR-balanced residual-strength
modulation. To improve multi-view feedback with a compact active-view set,
DReSG uses coverage-aware active views and conflict-filtered color updates. This
process converts expressive diffusion proposals into appearance updates that are
progressively absorbed by a shared renderable 3D scene.

In summary, our contributions are:

\begin{itemize}
  \item We formulate reference-guided 3DGS stylization as render-relative
  diffusion residual feedback,
  converting attention-guided diffusion proposals into render-relative targets
  that are progressively fitted into a shared Gaussian scene.

  \item We design a controllable residual-target and multi-view feedback
  procedure with RGB-logit target scaling, SNR-balanced residual modulation, and
  coverage-aware active views.

  \item Extensive comparisons and ablations against representative
  VGG-feature-based, CLIP-guided, and diffusion-based 3D stylization methods
  demonstrate a better balance among reference-guided stylization, content
  preservation, and cross-view stability.
\end{itemize}


\begin{figure*}[t]
  \centering
  \includegraphics[width=\textwidth]{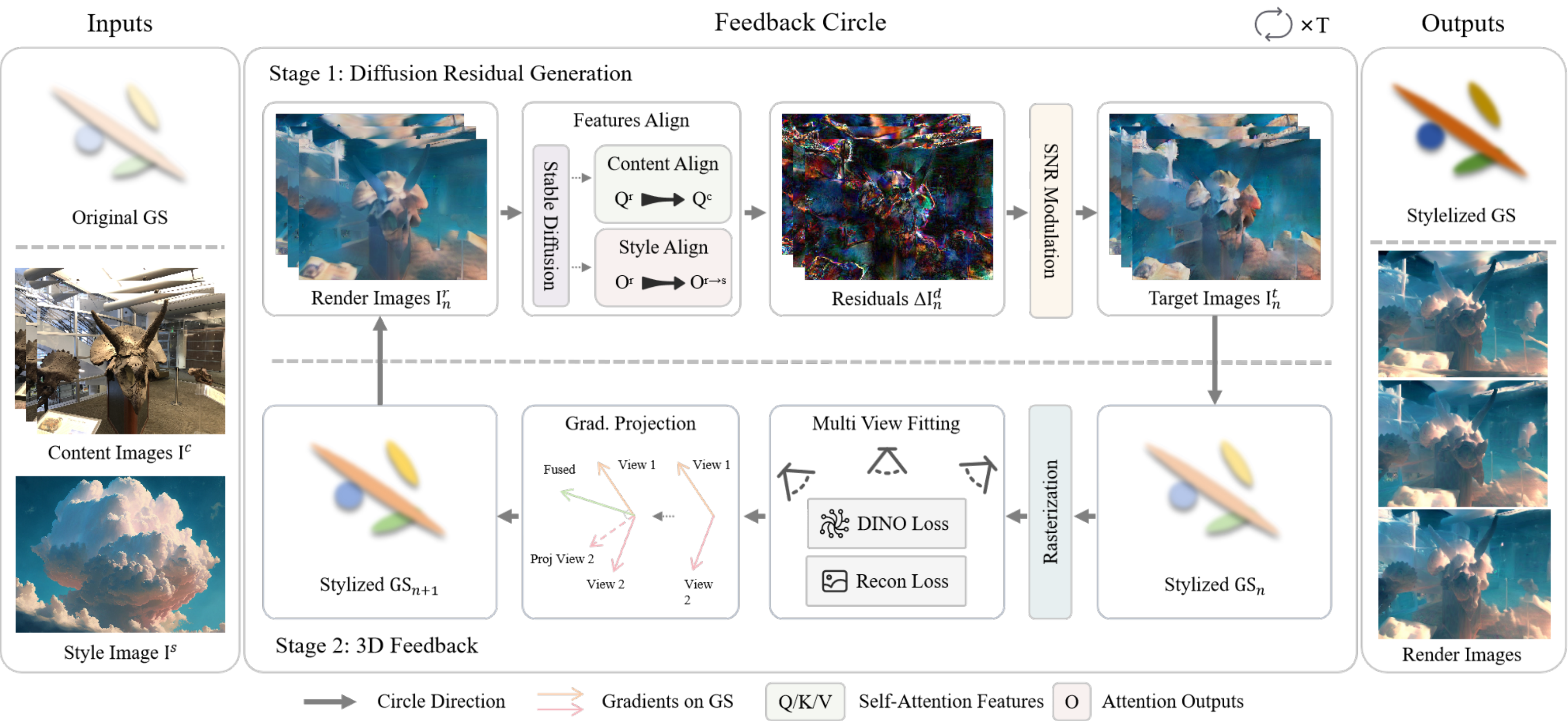}
  \caption{Overview of DReSG. Given a scene represented by 3DGS, its content renderings, and a reference style image, DReSG generates attention-guided diffusion proposals, constructs render-relative residual targets, and fits them back to the scene through multi-view Gaussian rendering.}
  \label{fig:method-overview}
\end{figure*}

\section{Related Work}
\label{sec:related-work}

\subsection{2D Reference-Guided Stylization}

2D reference-guided stylization studies how to transfer the appearance of a
style reference to an input image~\cite{Gatys2016Style,Johnson2016Perceptual,
Huang2017AdaIN,Li2017WCT}.
\historyadd{Classical neural style transfer represents style with Gram-matrix
correlations in VGG features and separates content and style in a deep feature
space. Subsequent methods improve efficiency and generality through perceptual
objectives, adaptive normalization, or feature transforms.}
\polish{More recently,} diffusion-based image stylization and
reference-guided generation use pretrained generative priors to synthesize
content-preserving images that follow a style or reference image
~\cite{Rombach2022LDM,Zhang2023InST,Ye2023IPAdapter,Wang2024InstantStyle}.
These methods commonly rely on latent diffusion, attention control, or
reference-image conditioning to produce more concrete color, stroke, and local
appearance changes than VGG-feature objectives alone.

\subsection{3D Scene Stylization}

3D scene stylization transfers the appearance of a reference image to a
renderable 3D representation, \polish{requiring the stylized appearance to remain
consistent under camera motion}. Existing approaches mainly build on two scene
representations. \historyadd{Neural Radiance Fields (NeRFs) represent scenes as
continuous neural radiance functions and render images through volumetric
integration~\cite{Mildenhall2020NeRF}.} NeRF-based stylization methods lift 2D style
objectives to 3D by optimizing implicit radiance fields with rendered-view
losses~\cite{Mildenhall2020NeRF,Nguyen2022SNeRF,Zhang2022ARF,Liu2023StyleRF,
Fujiwara2024StyleNeRF2NeRF}.
\historyadd{Later work improves view consistency by constructing stylized
observations or correspondences before or during 3D optimization
~\cite{Huang2021StylizeNovelViews,Huang2022StylizedNeRF,
Ibrahimli2024MuVieCAST,Zhu2025ReStyle3D}. These methods improve stylized novel
views, but their optimization and rendering costs are less aligned with the
explicit splatting representation used in 3DGS.}

Recent 3DGS stylization methods can be grouped by their supervision form.
VGG-feature
losses and patch-level matching provide stable Gaussian appearance
optimization~\cite{Liu2024StyleGaussian,Galerne2025SGSST,Liu2025ABCGS}, while
texture transfer and geometry-aware constraints improve controllability and
structure preservation~\cite{Liu2026GT2GS}. CLIP or prompt guidance introduces
more flexible semantic control~\cite{Howil2025CLIPGaussian}, and
diffusion-related guidance or stylized distillation strengthens reference-style
proposals~\cite{Yang2025FantasyStyle}. However, these supervision forms still
leave a gap between strong style signals and persistent scene-level appearance:
feature losses are indirect, prompt guidance is global, and diffusion-generated
stylized views can drift if their view-wise details are not absorbed by a shared
3D representation. DReSG addresses this gap with diffusion-derived,
render-relative residual targets that are progressively fitted into the shared
Gaussian scene.

\subsection{Diffusion-Guided 3D Generation and Editing}

Diffusion priors have been widely used for 3D generation and editing
~\cite{Poole2023DreamFusion,Shi2024MVDream,
Haque2023InstructNeRF2NeRF,Chen2024DGE}.
\historyadd{Existing methods broadly follow two routes. One uses a pretrained
2D diffusion model as a prior for 3D optimization, optimizing NeRF, mesh, or
Gaussian representations through score-based objectives or multi-view
diffusion priors~\cite{Poole2023DreamFusion,Wang2023SJC,Lin2023Magic3D,
Wang2023ProlificDreamer,Shi2024MVDream}. The other edits existing scenes by
applying instruction- or prompt-guided image edits to rendered views and
propagating those edits back to a 3D scene with multi-view constraints,
scene-level optimization, or consistency regularization
~\cite{Haque2023InstructNeRF2NeRF,Wang2024GaussianEditor,Wu2024GaussCtrl,
Wang2024VcEdit,Chen2024DGE}. These approaches demonstrate the
strength of diffusion priors as image-level guidance for 3D content generation
and editing.}
\polish{Compared with general diffusion-guided generation or editing,
reference-guided 3DGS stylization requires injecting reference appearance into a
reconstructed scene while preserving structure and cross-view persistence.} Score-based
optimization can produce timestep-dependent gradients without an inspectable
render-space target, whereas per-view edits may leave style changes in
independent images. DReSG instead uses diffusion to form observable
proposal-render residuals that are fitted by the shared Gaussian scene.


\section{Preliminaries}
\label{sec:preliminaries}

This section reviews \polish{the two sets of notation} used by DReSG: Gaussian
rendering and diffusion scheduler quantities. 3D Gaussian Splatting represents a
scene as a set of differentiable Gaussian primitives and renders images by
projecting visible Gaussians to the image plane followed by front-to-back alpha
compositing~\cite{Kerbl2023Gaussian}. Let
\(\mathcal{G}=\{(\mu_i,\mathbf{s}_i,\mathbf{q}_i,\alpha_i,
\mathbf{c}_i)\}_{i=1}^{N_{\mathcal G}}\) denote a Gaussian scene, where \(\mu_i\),
\(\mathbf{s}_i\), \(\mathbf{q}_i\), \(\alpha_i\), and \(\mathbf{c}_i\) are the
mean, scale, rotation, opacity, and color attributes of the \(i\)-th Gaussian.
Given a camera \(v\), the rendered color at pixel \(p\) is
\begin{equation}
  \mathcal{R}(\mathcal{G}_{n},v)[p]
  =
  \sum_{i\in\mathcal{N}_{p}}
  T_{i,p}\alpha_{i,p}\mathbf{c}_{i,p},
  \quad
  T_{i,p}=\prod_{j<i}(1-\alpha_{j,p}),
  \label{eq:gs-render}
\end{equation}
where \(\mathcal{N}_{p}\) is the depth-ordered set of Gaussians contributing to
pixel \(p\), \(\alpha_{i,p}\) is the projected opacity, \(T_{i,p}\) is the
accumulated transmittance before Gaussian \(i\), and \(\mathbf{c}_{i,p}\) is
the projected color. \polish{In the DReSG feedback loop, \(\mathcal{G}_n\) denotes
the current Gaussian scene at stage \(n\), and the rendered image from view \(v\)
is defined as}
\begin{equation}
  \mathbf{I}_{n}^{r,v} = \mathcal{R}(\mathcal{G}_{n}, v)\polish{.}
  \label{eq:render}
\end{equation}
Superscripts \(r\), \(c\), \(s\), and \(t\) denote the current render,
content-guidance image, style
image, and residual fitting target, respectively, while \(d\) denotes
proposal-render residuals. Throughout the paper, \(n\) indexes the outer DReSG
feedback stage and \(k\) indexes a selected diffusion scheduler state.

Diffusion models define a sequence of scheduler states that trade off signal and
noise during image generation~\cite{Ho2020DDPM,Rombach2022LDM}.
\polish{Under the standard forward-process notation, the noisy latent
corresponding to a clean latent \(\mathbf{z}_{0}\) at scheduler timestep \(k\) is
written as}
\begin{equation}
  \mathbf{z}_{k}
  =
  \sqrt{\bar{\alpha}_{k}}\mathbf{z}_{0}
  +
  \sqrt{1-\bar{\alpha}_{k}}\epsilon,
  \quad
  \bar{\alpha}_{k}=\prod_{i=1}^{k}\alpha_i .
  \label{eq:diffusion-forward}
\end{equation}
The corresponding signal-to-noise ratio is
\begin{equation}
  \mathrm{SNR}(k)=
  \frac{\bar{\alpha}_{k}}{1-\bar{\alpha}_{k}} .
  \label{eq:diffusion-snr}
\end{equation}
In DReSG, \(\bar{\alpha}_{k}\) and \(\mathrm{SNR}(k)\) are used only to describe
the scheduler state and to define the residual-strength modulation in
Section~\ref{sec:residual-targets}; the Gaussian scene parameters are not treated
as diffusion variables.

\section{Method}
\label{sec:method}

DReSG formulates stylized Gaussian splatting as a closed-loop residual feedback
problem grounded in rendered views. Given a scene represented by 3DGS and a
reference style image, DReSG repeatedly renders active views from the current
Gaussian scene, obtains attention-guided latent diffusion proposals from a
frozen diffusion model, converts proposal-render differences into
render-relative residual targets, and fits those targets into a shared Gaussian
scene. Fig.~\ref{fig:method-overview} summarizes this pipeline. The following
subsections describe the three main steps of DReSG: attention-guided proposal
residual generation, SNR-modulated residual target construction, and multi-view
Gaussian feedback. \rev{Algorithm~\ref{alg:residual-feedback} makes the alternating
latent and Gaussian optimization schedule explicit.}

\begin{algorithm}[t]
  \revcolor
  \caption{\rev{DReSG residual-feedback optimization.}}
  \label{alg:residual-feedback}
  \footnotesize
  \begin{algorithmic}[1]
    \renewcommand{\algorithmicrequire}{\textbf{Input:}}
    \renewcommand{\algorithmicensure}{\textbf{Output:}}
    \Require Base scene $\mathcal{G}$, active views $\mathcal{V}$, content views $\{\mathbf{I}^{c,v}\}$, style image $\mathbf{I}^{s}$, DDIM states $\{\tau_q\}_{q=1}^{N}$, feedback interval $S$, and fit length $M$
    \State $\mathbf{z}^{r,v}\gets\mathcal{E}_{\mathrm{VAE}}(\operatorname{Render}(\mathcal{G},v)),\quad\forall v\in\mathcal{V}$
    \For{$n\gets 1$ \algorithmicto{} $N/S$}
      \For{$j\gets 1$ \algorithmicto{} $S$}
        \State $k\gets\tau_{S(n-1)+j}$
        \State $\mathbf{z}^{r,v}\gets\operatorname{Adam}(\mathbf{z}^{r,v};\mathcal{E}^{\mathrm{attn},v}(k)),\quad\forall v\in\mathcal{V}$
      \EndFor
      \State $\mathbf{I}^{r,v}\gets\operatorname{Render}(\mathcal{G},v),\quad\Delta\mathbf{I}^{d,v}\gets\operatorname{Decode}(\mathbf{z}^{r,v})-\mathbf{I}^{r,v}$
      \State $\mathbf{I}^{t,v}\gets\operatorname{Target}_{k}(\mathbf{I}^{r,v},\Delta\mathbf{I}^{d,v})$ using Eq.~\ref{eq:residual-target}, $\forall v\in\mathcal{V}$
      \For{$m\gets 1$ \algorithmicto{} $M$}
        \State Compute $\mathcal{L}_{\mathrm{fit}}$ and view-wise color gradients.
        \State Project and fuse conflicting color gradients.
        \State $\mathcal{G}\gets\operatorname{Adam}(\mathcal{G};\mathcal{L}_{\mathrm{fit}})$
      \EndFor
      \State $\mathbf{z}^{r,v}\gets\mathcal{E}_{\mathrm{VAE}}(\operatorname{Render}(\mathcal{G},v)),\quad\forall v\in\mathcal{V}$
    \EndFor
    \Ensure Stylized Gaussian scene $\mathcal{G}$
  \end{algorithmic}
\end{algorithm}

\subsection{Attention-Guided Proposal Residuals}
\label{sec:attn-diffusion-residual}

An expressive style signal should be extracted without decoupling it from the
current 3D scene. VGG-style rendered-view losses are stable for optimization,
yet their supervision is often implicit in feature-space correlations or patch matches.
Diffusion models can propose richer reference-aware appearance changes, but a
full per-view stylized proposal may contain details that cannot be explained
consistently by a shared Gaussian scene. DReSG therefore uses diffusion only to
propose an appearance update for the current render\polish{; the resulting
proposal-render residual then guides subsequent 3D feedback.}

For each
active view \(v\), a frozen latent diffusion model with attention-guided latent
optimization receives the current rendering \(\mathbf{I}_{n}^{r,v}\), the
original content view \(\mathbf{I}^{c,v}\), and the style image
\(\mathbf{I}^{s}\). \polish{The index \(n\) identifies the current 3D scene state,
whereas \(k\) identifies the diffusion feature-extraction state.} The current render is the
image being stylized, the content view provides layout guidance, and the style
image provides reference appearance. At each feedback stage, the render latent
is initialized from the VAE projection of the current render and optimized with
attention guidance at the scheduler state used by that stage.

DReSG obtains the residual by optimizing the render latent along selected
scheduler indices. Let \(\mathbf{z}_{k}^{r,v}\) denote the current render
latent state during this optimization at scheduler state \(k\),
\(\mathbf{z}^{c,v}\) the VAE latent of \(\mathbf{I}^{c,v}\), and
\(\mathbf{z}^{s}\) the VAE latent of \(\mathbf{I}^{s}\). We use the frozen U-Net
as a timestep-conditioned feature extractor: \(k\) selects the U-Net
features used for attention guidance and also provides the scheduler state for
residual-strength modulation. Let
\begin{equation}
  \begin{aligned}
  (\mathbf{Q}_{k}^{r,v},\mathbf{K}_{k}^{r,v},\mathbf{V}_{k}^{r,v})
  &=
  \Phi^{\mathrm{attn}}(\mathbf{z}_{k}^{r,v};k),\\[2pt]
  (\mathbf{Q}_{k}^{c,v},\mathbf{K}_{k}^{c,v},\mathbf{V}_{k}^{c,v})
  &=
  \Phi^{\mathrm{attn}}(\mathbf{z}^{c,v};k),\\[2pt]
  (\mathbf{Q}_{k}^{s},\mathbf{K}_{k}^{s},\mathbf{V}_{k}^{s})
  &=
  \Phi^{\mathrm{attn}}(\mathbf{z}^{s};k).
  \end{aligned}
  \label{eq:attn-features}
\end{equation}
\polish{The operator \(\Phi^{\mathrm{attn}}\) collects self-attention features
from selected U-Net layers at timestep \(k\) into compact Q/K/V feature tensors.}
We use the content
query for layout preservation and the style keys/values for reference
appearance guidance; content and style features serve as fixed references
during render-latent optimization. Content
guidance matches the current render query to the content
query,
\begin{equation}
  \mathcal{E}^{c,v}(k)
  =
  \left\|
    \mathbf{Q}_{k}^{r,v}
    - \mathbf{Q}_{k}^{c,v}
  \right\|_{1}.
  \label{eq:content-query-loss}
\end{equation}
For style guidance, we compare the current self-attention output with an output
formed by using the render query to attend to the style keys and values:
\begin{equation}
  \begin{aligned}
  \mathbf{O}_{k}^{r,v}
  &=
  \mathrm{Attn}(\mathbf{Q}_{k}^{r,v},
  \mathbf{K}_{k}^{r,v},\mathbf{V}_{k}^{r,v}),\\[2pt]
  \mathbf{O}_{k}^{r\rightarrow s,v}
  &=
  \mathrm{Attn}(\mathbf{Q}_{k}^{r,v},
  \mathbf{K}_{k}^{s},\mathbf{V}_{k}^{s}).
  \end{aligned}
  \label{eq:style-attention}
\end{equation}
\polish{The operator \(\mathrm{Attn}(\cdot)\) denotes the standard
attention-output operator of the frozen U-Net self-attention blocks.} The style loss is
\begin{equation}
  \mathcal{E}^{s,v}(k)
  =
  \left\|
    \mathbf{O}_{k}^{r,v}
    - \mathbf{O}_{k}^{r\rightarrow s,v}
  \right\|_{1}.
  \label{eq:style-output-loss}
\end{equation}
This \polish{formulation} is inspired by \polish{the self-attention
feature-transfer objective of Zhou et al.}~\cite{Zhou2025AttentionDistillation}.
DReSG then uses the optimized render latent to construct proposal-render
residuals for Gaussian feedback. The attention guidance energy is
\begin{equation}
  \mathcal{E}^{\mathrm{attn},v}(k)
  =
  \mathcal{E}^{s,v}(k)+w_c\mathcal{E}^{c,v}(k).
  \label{eq:attn-guidance-energy}
\end{equation}
\polish{The scalar \(w_c\) controls content preservation.} At each feedback stage, DReSG
updates the render latent for multiple Adam steps to reduce
\(\mathcal{E}^{\mathrm{attn},v}\). We denote the resulting optimized proposal
latent as \(\mathbf{z}_{k}^{r,v,*}\). The proposal-render residual is obtained by
decoding the optimized latent and subtracting the current render:
\begin{equation}
  \Delta\mathbf{I}_{n}^{d,v}(k)
  =
  \mathcal{D}_{\mathrm{VAE}}(\mathbf{z}_{k}^{r,v,*})
  -
  \mathbf{I}_{n}^{r,v}.
  \label{eq:diffusion-residual}
\end{equation}

\subsection{SNR-Modulated Residual Targets}
\label{sec:residual-targets}

\polish{DReSG converts each proposal-render residual into a fitting target whose
strength is explicitly controlled.} A weak residual may fail to transfer
recognizable style cues, whereas an overly amplified residual can saturate
colors or force unstable local appearance changes into the Gaussian scene.
\polish{To account for scheduler-dependent proposal behavior, DReSG constructs a
bounded fitting target by modulating residual strength and amplifying the
residual in RGB-logit space.} Given the current render \(\mathbf{I}_{n}^{r,v}\)
and the residual
\(\Delta\mathbf{I}_{n}^{d,v}(k)\), the corresponding unscaled diffusion proposal
is \(\mathbf{I}_{n}^{r,v}+\Delta\mathbf{I}_{n}^{d,v}(k)\).

\begin{figure}[t]
  \centering
  \includegraphics[width=0.92\linewidth]{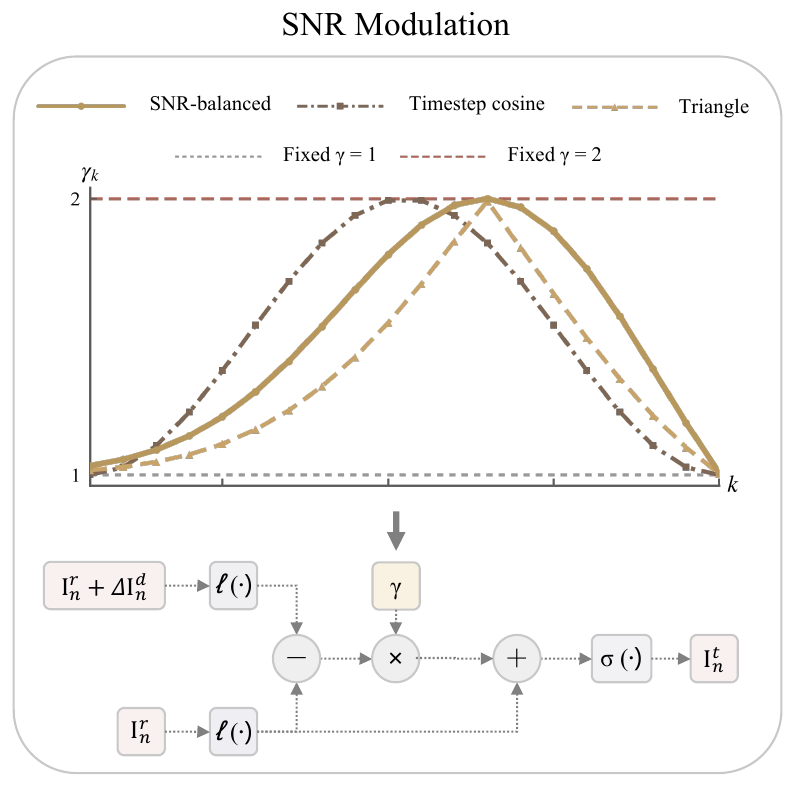}
  \caption{\rev{Residual-strength modulation and RGB-logit target construction. Top: the SNR-balanced schedule and four alternatives evaluated in Section~\ref{sec:ablation}. Bottom: the RGB-logit construction of the stage target from the current render and diffusion proposal.}}
  \label{fig:snr-modulation}
\end{figure}

\polish{We define the channel-wise RGB logit transform, used only for target
construction, as
\(\ell(\mathbf{I})=\log\frac{\mathbf{I}}{1-\mathbf{I}}\).}
\rev{We parameterize the residual scale as \(\gamma_k=1+p_k\), with
\(p_k\in[0,1]\), preserving the unamplified proposal as the baseline while
allowing only bounded extrapolation. Prior diffusion studies report that guidance
benefits and representation quality vary across scheduler states and are
strongest over intermediate regimes~\cite{kynkaanniemi2024applying,
li2026understanding}. We therefore use a state-dependent unimodal schedule
rather than a constant gain. Under the variance-preserving scheduler,
\(\bar{\alpha}_k\) and \(1-\bar{\alpha}_k\) are the signal and noise power
fractions; their peak-normalized product defines the SNR-balanced schedule:}
\begin{equation}
  \revcolor
  \gamma_k
  =
  1+p_k^{\mathrm{snr}}
  =
  1+4\bar{\alpha}_{k}(1-\bar{\alpha}_{k})
  =
  1+\frac{4\,\mathrm{SNR}(k)}
          {(1+\mathrm{SNR}(k))^2}.
  \label{eq:snr-residual-weight}
\end{equation}
\rev{The resulting scale is smooth and bounded in \([1,2]\), peaks at
\(\mathrm{SNR}(k)=1\), and is symmetric under
\(\mathrm{SNR}\mapsto1/\mathrm{SNR}\). It returns to \(\gamma_k=1\) when
either signal or noise dominates, retaining the proposal while reducing only
the extra amplification. As the schedule depends only on the scheduler state,
equal-SNR states receive equal scales regardless of schedule discretization.
Fig.~\ref{fig:snr-modulation} visualizes this modulation profile and the
RGB-logit target construction; the alternative profiles in its top panel are
evaluated in Section~\ref{sec:ablation}.}

The final fitting target extrapolates from the current render toward the
diffusion proposal in RGB-logit space:
\begin{equation}
  \mathbf{I}_{n}^{t,v}(k) =
  \sigma\!\left(
  \ell(\mathbf{I}_{n}^{r,v})
  +
  \gamma_k
  \left[
  \ell(\mathbf{I}_{n}^{r,v}+\Delta\mathbf{I}_{n}^{d,v}(k))
  -
  \ell(\mathbf{I}_{n}^{r,v})
  \right]
  \right).
  \label{eq:residual-target}
\end{equation}

The target is defined relative to the current render rather than a fixed
stylized image. As optimization proceeds across feedback stages, style changes
that have already been absorbed by the 3D scene produce smaller residuals in
later stages, whereas view-specific proposal details must be reproduced through
the shared scene or appear as a remaining fitting discrepancy. RGB-logit scaling
only defines the space in which these residual targets are amplified, making it
a bounded target-construction step rather than a separate source of 3D
structure.

\subsection{Multi-View Gaussian Feedback}
\label{sec:multi-view-feedback}

Proposal targets from multiple views must be absorbed by one shared Gaussian
scene to become persistent 3D appearance. Even when each view has a reasonable
fitting target, different targets may supervise overlapping Gaussians with
inconsistent color updates, and using all available views would make proposal
generation unnecessarily expensive. DReSG therefore selects a compact set of
active views, fits their targets through differentiable multi-view Gaussian
rendering, and filters conflicting color updates before applying them to the
shared scene. It does not introduce a new Gaussian representation; it optimizes
the Gaussian color attributes \(\mathbf{c}\) and bounded mean, scale, and
rotation offsets, while keeping opacity frozen.

DReSG selects active views offline using visibility-based Gaussian coverage.
Let \(w_{v,i}\) measure how strongly Gaussian \(i\) is visibly observed in
candidate view \(v\). We compute this value from projected alpha and opacity
contributions after excluding samples whose depth is inconsistent with the
rendered depth. With minimum support \(w_{\min}\) and target fraction \(\rho\),
we compute \(m_i^\star=\max_v w_{v,i}\), keep the target-visible set
\(\mathcal{T}=\{i\mid m_i^\star\geq w_{\min}\}\), and define the per-Gaussian
target support
\begin{equation}
  q_i = \max(\rho\,m_i^\star,w_{\min}) .
  \label{eq:target-support}
\end{equation}
For a selected view set \(\mathcal{A}\), we define the capped coverage of
Gaussian \(i\) as
\begin{equation}
  c_i(\mathcal{A})
  =
  \min\!\left(
  \frac{\max_{u\in\mathcal{A}}w_{u,i}}{q_i},
  1
  \right).
  \label{eq:view-coverage}
\end{equation}
\rev{For the empty initial set, we define \(c_i(\emptyset)=0\).}
\rev{Starting from an empty set, DReSG greedily adds the candidate view that
maximizes the total increase of capped coverage:}
\begin{equation}
  v^\star
  =
  \arg\max_{v\in\mathcal{C}\setminus\mathcal{A}}
  \sum_{i\in\mathcal{T}}
  \left[
  c_i(\mathcal{A}\cup\{v\})-c_i(\mathcal{A})
  \right].
  \label{eq:greedy-view-selection}
\end{equation}
\rev{The procedure terminates when the prescribed coverage threshold is reached
or the marginal gain falls below its threshold.} \polish{The selected view set
remains fixed during optimization and determines only where proposal targets
are generated.} \rev{The resulting fixed
active-view set is denoted by \(\mathcal{V}\).}

Given the selected active views, DReSG uses a standard per-view image-space
fitting objective for residual feedback:
\begin{equation}
  \mathcal{L}_{\mathrm{fit}}^{v}
  =
  \mathcal{L}_{1}^{v}
  +\lambda_{\mathrm{ssim}}\mathcal{L}_{\mathrm{DSSIM}}^{v}
  +\lambda_{\mathrm{tv}}\mathcal{L}_{\mathrm{tv}}^{v}
  +\lambda_{\mathrm{dino}}\mathcal{L}_{\mathrm{DINO}}^{v}.
  \label{eq:fit-loss}
\end{equation}
Each per-view loss is averaged over image pixels, and the DINO feature loss uses
the base render as its content reference. Geometry is updated only through small
offsets from the base Gaussian means, scales, and rotations, and these offsets
are projected back to preset ranges after each update. \polish{This objective
grounds independent diffusion proposals in 3D: every target must be reproduced
by the same Gaussian scene under differentiable rendering.}

\rev{Residual feedback from different active views may disagree on overlapping
Gaussian regions. DReSG applies color-gradient projection to conflicting view
updates during appearance-gradient fusion.}
\polish{Each per-view objective in Eq.~\ref{eq:fit-loss} induces a gradient on
the shared Gaussian color attributes \(\mathbf{c}\):}
\begin{equation}
  \mathbf{g}_{v}^{c}=\nabla_{\mathbf{c}}\mathcal{L}_{\mathrm{fit}}^{v}.
  \label{eq:view-gradient}
\end{equation}
If two views \(u\) and \(v\) propose opposing updates,
\begin{equation}
  (\mathbf{g}_{u}^{c})^{\top}\mathbf{g}_{v}^{c}<0,
  \label{eq:gradient-conflict}
\end{equation}
we use color-gradient projection, inspired by gradient surgery~\cite{Yu2020PCGrad},
to remove the directly conflicting component, using \(\epsilon>0\) for
numerical stability:
\begin{equation}
  \tilde{\mathbf{g}}_{u}^{c}
  =
  \mathbf{g}_{u}^{c}
  -
  \frac{(\mathbf{g}_{u}^{c})^{\top}\mathbf{g}_{v}^{c}}
       {\|\mathbf{g}_{v}^{c}\|_2^2+\epsilon}
  \mathbf{g}_{v}^{c}.
  \label{eq:gradient-projection}
\end{equation}
\polish{The adjusted gradients are then averaged to obtain the final
color-update direction:}
\begin{equation}
  \bar{\mathbf{g}}^{c} =
  \frac{1}{|\mathcal{V}|}
  \sum_{v\in\mathcal{V}}\tilde{\mathbf{g}}_{v}^{c}.
  \label{eq:fused-gradient}
\end{equation}
For multiple active views, DReSG processes gradients in the deterministic
active-view order. For each view gradient, it sequentially checks conflicts
against the other active-view gradients and updates the current gradient with
Eq.~\ref{eq:gradient-projection} whenever the inner product is negative. After
all view gradients are processed, the projected gradients are averaged as in
Eq.~\ref{eq:fused-gradient}. Bounded geometry gradients are fused by a simple
mean and constrained by their offset ranges, while opacity receives no update.
\rev{Thus color-gradient projection suppresses directly conflicting, view-specific
local updates during fusion, helping reduce view-dependent artifacts and local
style fragmentation without changing the residual targets or feedback schedule.}

After each 3D feedback stage, the updated scene is rendered again and \rev{re-encoded
to initialize the render latent for the next stage}, so each residual describes
the remaining stylization discrepancy of the current fitted scene rather than a
fixed target generated from the
initial rendering. A standard affine color-transfer fitting step can be applied
at the end~\cite{Zhang2022ARF,Liu2025ABCGS,Liu2026GT2GS}, yielding a stylized
Gaussian scene renderable along novel camera paths within the reconstructed
scene.


\begin{figure*}[t]
  \centering
  \includegraphics[width=\textwidth]{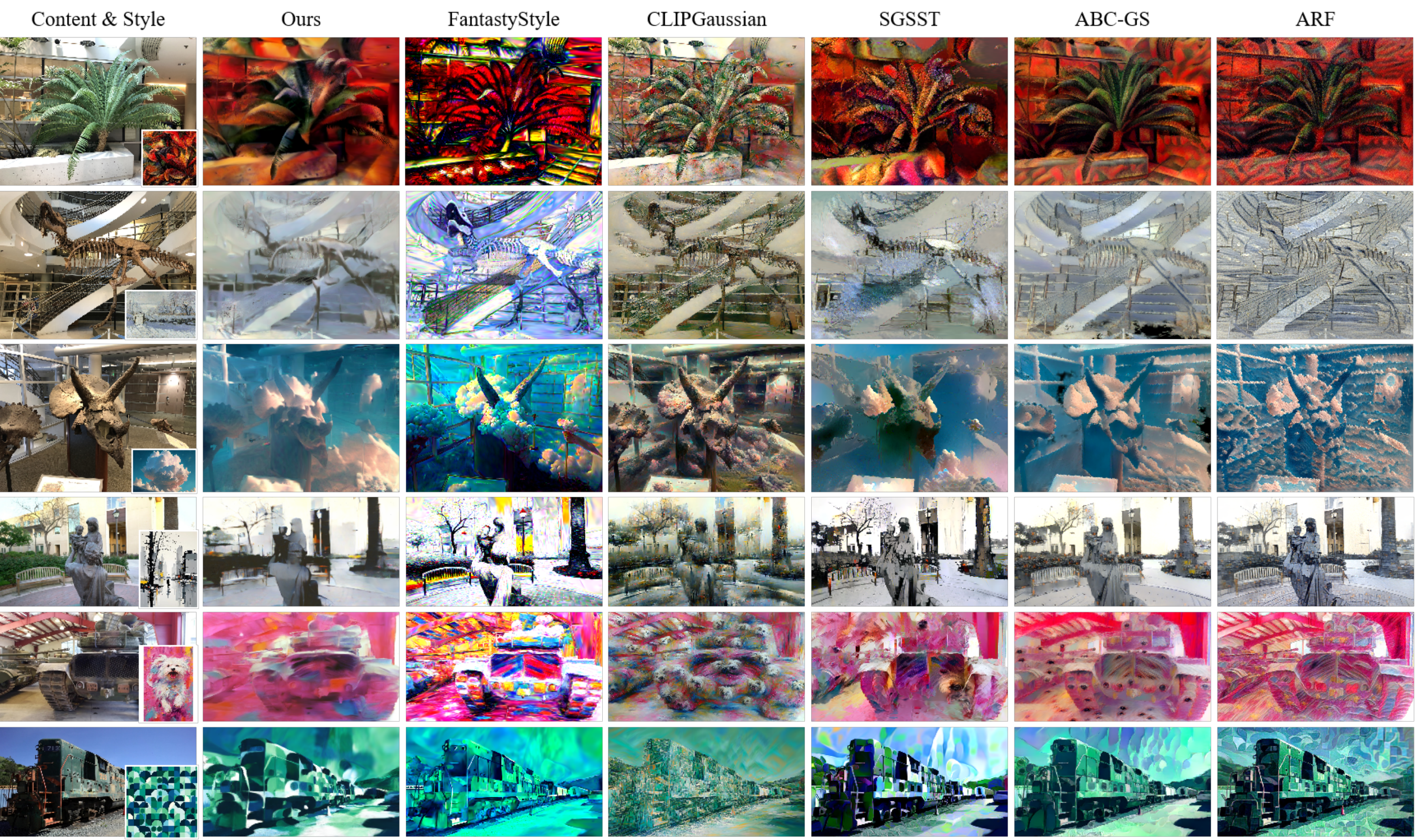}
  \caption{Main qualitative comparison on LLFF and Tanks and Temples scenes. Rows are selected to cover reference-style properties discussed in Sec.~\ref{sec:introduction}, including directional strokes, coherent contours, region-wise color treatment, and structured style details. All methods are rendered from matched camera poses.}
  \label{fig:main-qualitative}
\end{figure*}

\begin{table*}[t!]
  \centering
  \definecolor{RankFirstBg}{RGB}{255,223,128} 
  \definecolor{RankSecondBg}{RGB}{220,224,229} 
  \definecolor{RankThirdBg}{RGB}{218,168,112} 
  \providecommand{\rankfirst}[1]{\cellcolor{RankFirstBg}#1}
  \providecommand{\ranksecond}[1]{\cellcolor{RankSecondBg}#1}
  \providecommand{\rankthird}[1]{\cellcolor{RankThirdBg}#1}
  \caption{Main quantitative comparison averaged over evaluated scenes and styles. Temporal columns report RAFT-aligned temporal drift at short and long strides; memory is measured in GB.}
  \label{tab:main-quantitative}
  \resizebox{\linewidth}{!}{
    \begin{tabular}{lcccccccccc}
      \toprule
      Method &
      \multicolumn{1}{c}{Style} &
      \multicolumn{1}{c}{Content} &
      \multicolumn{2}{c}{Short-term} &
      \multicolumn{2}{c}{Long-term} &
      \multicolumn{4}{c}{Efficiency} \\
      \cmidrule(lr){2-2}
      \cmidrule(lr){3-3}
      \cmidrule(lr){4-5}
      \cmidrule(lr){6-7}
      \cmidrule(lr){8-11}
      & CLIP-S \(\uparrow\) & DINO-C \(\uparrow\) &
      ST-LPIPS \(\downarrow\) & ST-RMSE \(\downarrow\) &
      LT-LPIPS \(\downarrow\) & LT-RMSE \(\downarrow\) &
      Opt. Time (min) \(\downarrow\) & Opt. Mem. (GB) \(\downarrow\) &
      Infer Mem. (GB) \(\downarrow\) & FPS \(\uparrow\) \\
      \midrule
      Ours & \rankthird{0.6773} & \rankfirst{0.5803} &
      \rankfirst{0.0648} & \rankfirst{0.0321} &
      \rankfirst{0.1045} & \rankfirst{0.0501} &
      \rankthird{7.80} & 15.65 & \rankfirst{1.80} &
      \rankfirst{248.88} \\
      FantasyStyle & 0.6600 & \rankthird{0.4260} &
      \ranksecond{0.0708} & 0.0890 &
      \ranksecond{0.1100} & 0.1342 &
      112.90 & 33.30 & 2.26 & 168.60 \\
      CLIPGaussian & \ranksecond{0.7251} & 0.4055 &
      0.0993 & 0.0841 &
      0.1292 & 0.1176 &
      13.14 & 5.40 & 2.25 & 211.03 \\
      SGSST & \rankfirst{0.7271} & 0.1836 &
      0.0865 & \rankthird{0.0680} &
      0.1172 & \rankthird{0.0986} &
      21.79 & \rankfirst{2.40} & 2.26 &
      \rankthird{212.06} \\
      ABC-GS & 0.6282 & \ranksecond{0.4438} &
      \rankthird{0.0832} & \ranksecond{0.0515} &
      \rankthird{0.1140} & \ranksecond{0.0723} &
      \rankfirst{1.06} & \rankthird{5.08} & \rankthird{2.20} &
      \ranksecond{239.52} \\
      ARF & 0.6674 & 0.3126 &
      0.1117 & 0.0766 &
      0.1610 & 0.1090 &
      \ranksecond{3.32} & \ranksecond{4.45} &
      \ranksecond{1.82} & 10.59 \\
      \bottomrule
    \end{tabular}
  }
\end{table*}

\begin{figure*}[!t]
  \centering
  \includegraphics[width=\textwidth]{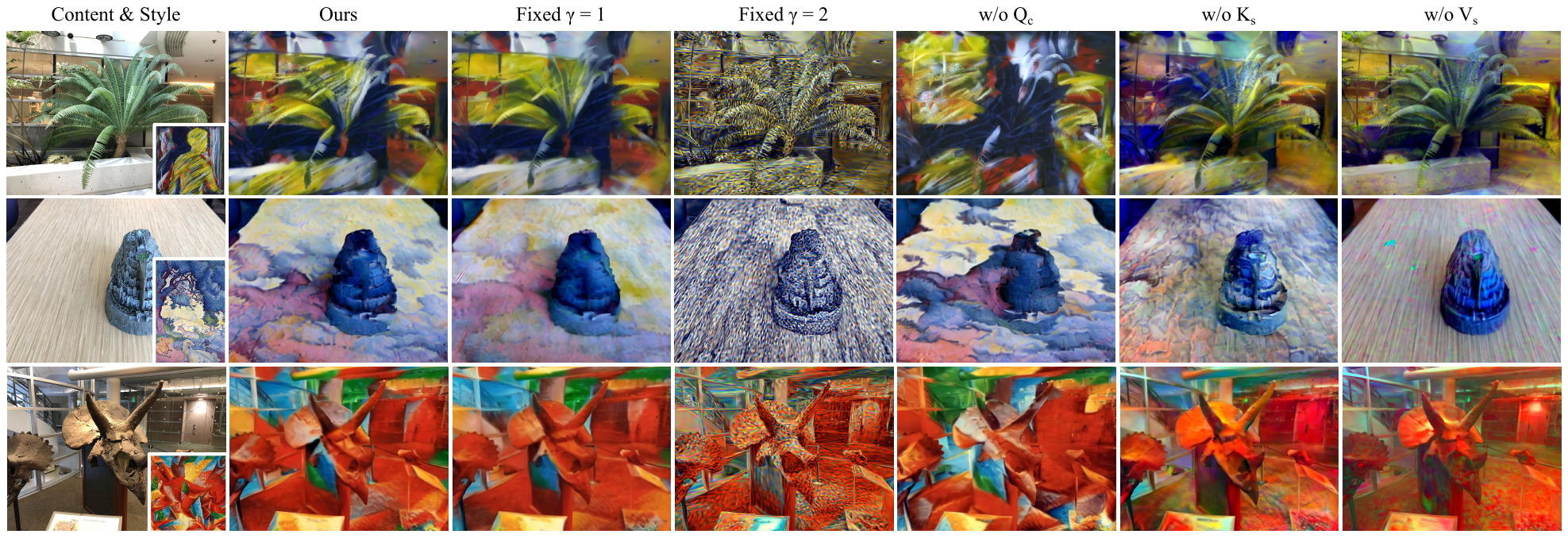}
  \caption{\rev{Attention-guided residual construction and representative fixed-schedule ablations. The schedule columns show SNR-balanced and the two fixed endpoints; all five residual-strength schedules are compared quantitatively in Table~\ref{tab:schedule-ablation}.}}
  \label{fig:residual-target-ablation}
\end{figure*}

\section{Experiments}
\label{sec:experiments}

We evaluate DReSG for reference-guided 3DGS stylization with quantitative
results on LLFF~\cite{Mildenhall2019LLFF} and qualitative comparisons on both
LLFF and Tanks and Temples~\cite{Knapitsch2017Tanks}. The main quantitative
benchmark contains 48 LLFF scene-style pairs. We report main comparisons,
ablations, and a user study. In the tables, colored cells indicate the top
three entries per main-comparison column and the top two per ablation column.

\subsection{Metrics}
\label{sec:metrics}

\polish{We report six quantitative metrics that separately assess style
alignment, content preservation, and cross-view consistency. CLIP-S measures
CLIP~\cite{Radford2021CLIP} feature similarity between the stylized render and
the reference style image; CLIP features capture high-level image appearance
and reference-style cues across visual domains. DINO-C measures
DINO~\cite{Caron2021DINO} feature similarity between the stylized render and
the corresponding content image at the same camera pose; DINO features
emphasize object- and scene-level correspondence rather than low-level color
matching, providing a content-preservation measure.}
We compute CLIP-S with a pretrained CLIP ViT-B/32 backbone and DINO-C with a
pretrained DINO ViT-S/16 backbone. Images are resized to \(224\times224\) and
normalized with the corresponding pretrained-model preprocessing. Scores are
averaged over all evaluated views, styles, and scenes. The main quantitative
evaluation uses 48 LLFF scene-style pairs. For each scene-style pair, all
methods are evaluated on the same
rendered camera views.
We denote short-term and long-term temporal metrics as ST and LT in tables.
In compact ablation tables, C-S and D-C abbreviate CLIP-S and DINO-C, while LP
and RM abbreviate LPIPS and RMSE.

Table~\ref{tab:main-quantitative} also reports optimization time, peak
optimization memory, peak inference memory, and inference FPS. All resource
measurements are taken on a single
NVIDIA H20 GPU at the same rendering resolution. Optimization time includes
diffusion proposal generation and residual fitting, but excludes the initial
base 3DGS reconstruction. Inference FPS and inference memory are measured on
the final stylized Gaussian scene without diffusion calls.

For temporal consistency, we use the Princeton-VL RAFT optical-flow
model~\cite{Teed2020RAFT} to align frame pairs. To avoid counting camera motion
as stylization flicker, optical flow is estimated on the original base renders
and then used to compare the corresponding stylized frames. Short-term metrics
use adjacent frames with gap \(1\), and long-term metrics use gap \(N_v/2\),
where \(N_v\) is the number of evaluated views. RAFT forward--backward
consistency and \polish{in-bounds} checks define valid regions. RMSE is averaged over
valid pixels; for LPIPS~\cite{Zhang2018LPIPS}, invalid pixels in the warped
stylized frame are replaced with target-frame pixels before computing the
full-image distance. Lower RAFT-aligned LPIPS and RMSE indicate less appearance
drift.

\subsection{Baselines}

We compare DReSG with representative 3D scene stylization methods covering the
main supervision and representation choices in this task. ARF~\cite{Zhang2022ARF}
is a NeRF-based rendered-view feature-loss baseline; SGSST~\cite{Galerne2025SGSST}
and ABC-GS~\cite{Liu2025ABCGS} are recent Gaussian stylization pipelines driven
by VGG-style feature losses and patch-level matching; CLIPGaussian~\cite{Howil2025CLIPGaussian}
evaluates CLIP-guided multimodal Gaussian stylization; and
FantasyStyle~\cite{Yang2025FantasyStyle} provides a diffusion-based 3D
stylization comparison. This set covers the principal alternatives to DReSG's
residual-feedback formulation: rendered-view feature optimization, patch-level
Gaussian fitting, CLIP guidance, and diffusion-guided 3D stylization.

Unless otherwise stated, all methods use the same input scene, reference style
image, base Gaussian reconstruction, and evaluation cameras. Gaussian-based
methods are initialized from the same base reconstruction for each scene, since
reconstruction defects can otherwise dominate perceived stylization quality.
\polish{ARF is evaluated with its method-specific NeRF checkpoint; this method is
not defined for Gaussian checkpoints.} We run baselines with official implementations
and released configurations when available, and otherwise use the authors'
default hyperparameters.

\subsection{Implementation Details}

DReSG starts from a Fast-PGSR all-view base built on FastGS acceleration and
PGSR reconstruction~\cite{Ren2025FastGS,Chen2024PGSR}. It optimizes Gaussian
color attributes \(\mathbf{c}\) and small offsets to means, scales, and
rotations, while opacity is frozen; geometry offsets are projected back to
preset ranges after each update. Active views are selected once before
optimization using the offline coverage strategy in
Section~\ref{sec:multi-view-feedback}. The default setting uses
\(w_{\min}=10^{-4}\), \(\rho=0.98\), coverage-stop ratio \(0.9999\), and
marginal-gain threshold \(0.001\). \rev{Selection begins from the empty set and
terminates when either prescribed stopping condition is met; it typically
retains approximately 20 LLFF views and 56--70 Tanks and Temples views.}
\rev{LLFF scenes are rendered at factor four, whereas Tanks and Temples scenes
use their native resolution. All renderings and reference images
are resized to \(448\times320\) before being passed to the diffusion model.}

\begin{figure*}[!t]
  \centering
  \includegraphics[width=\textwidth]{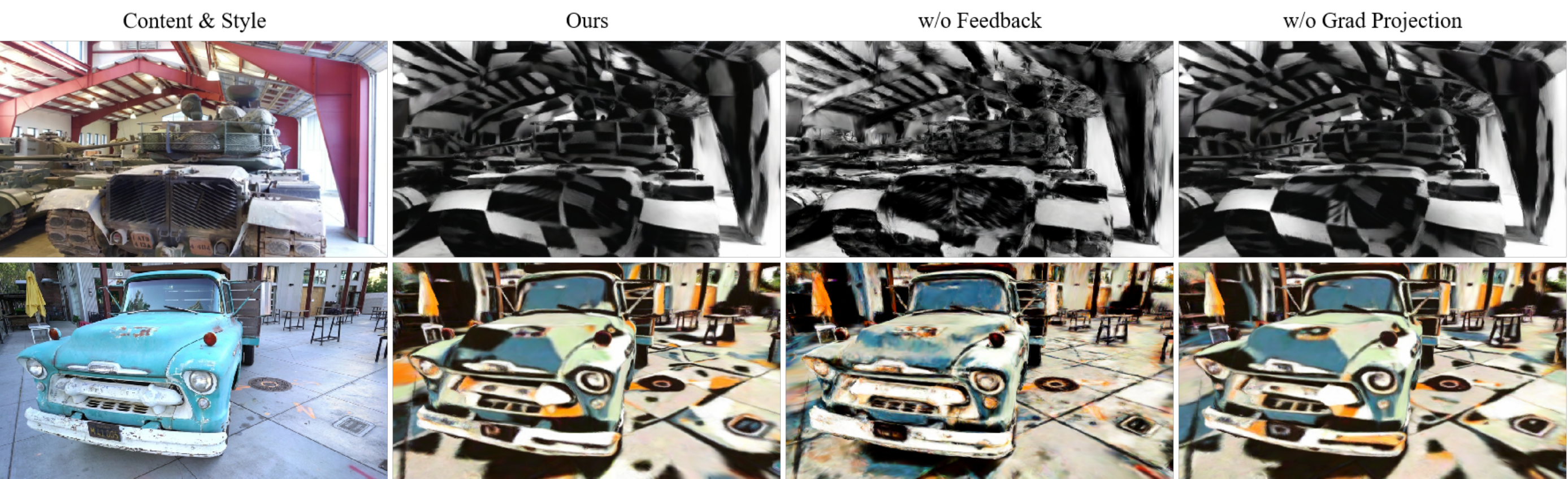}
  \caption{3D-grounded feedback ablation. We compare DReSG with variants that remove residual feedback or color-gradient projection under matched scenes, styles, and viewpoints.}
  \label{fig:feedback-ablation}
\end{figure*}

\begin{figure}[!t]
  \centering
  \includegraphics[width=\linewidth]{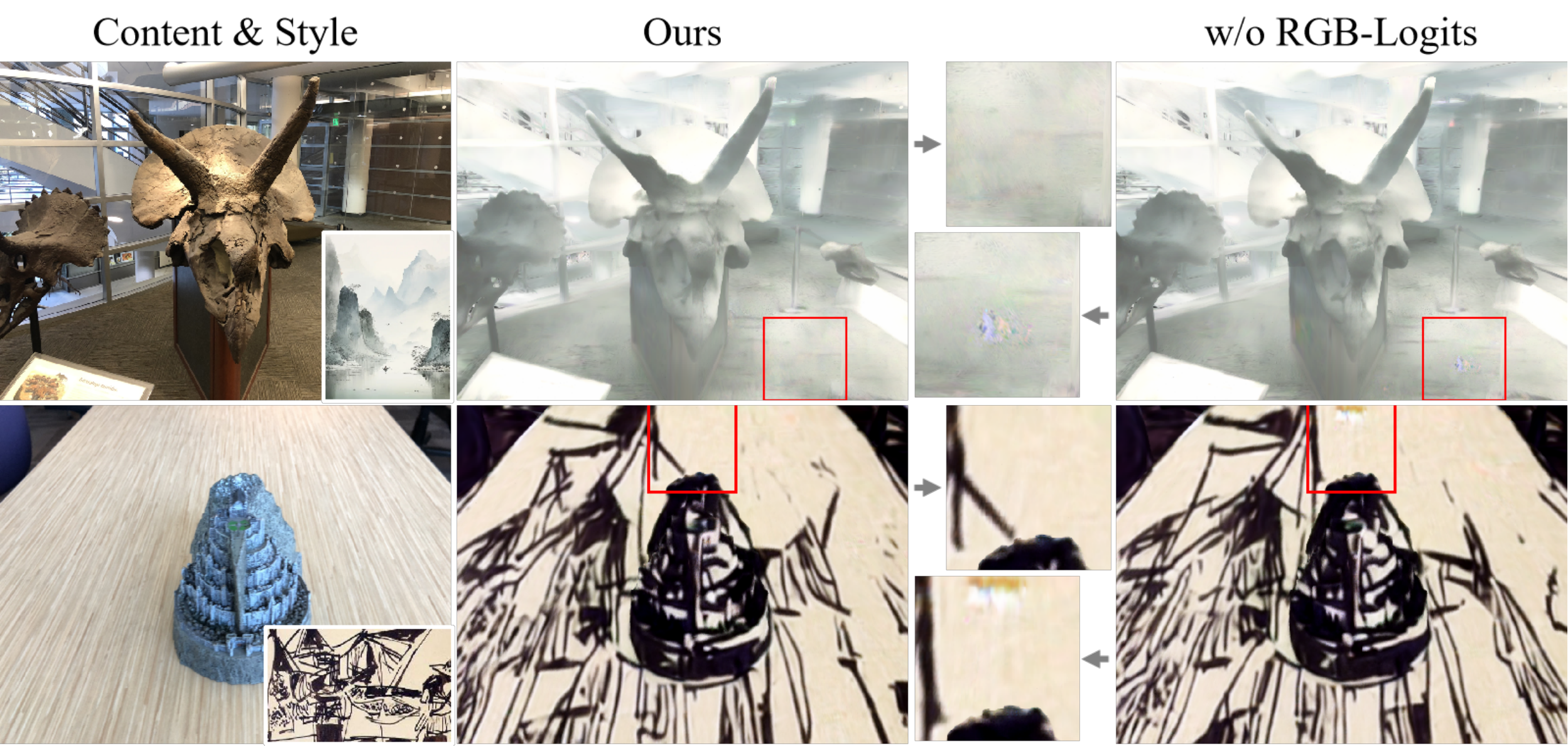}
  \caption{RGB-logit target scaling ablation. We compare DReSG with an image-space residual-scaling variant under matched scenes, styles, and viewpoints.}
  \label{fig:logit-ablation}
\end{figure}

At each feedback stage, we render the current 3D scene from the active views,
generate attention-guided diffusion proposals with Stable Diffusion v1.5 using
the attention energy in Eq.~\ref{eq:attn-guidance-energy}, compute
render-relative residual targets with RGB-logit scaling, and fit the scene to
those targets. This logit scaling is applied only during target construction.
The main setting uses 20 feedback stages, whose scheduler states are sampled
uniformly from a 200-step DDIM schedule. Attention features are extracted from
\rev{zero-indexed U-Net self-attention layers 10--15}. After all attention-guided
residual stages are complete, we run the post affine color-transfer fitting step
described above. The default configuration uses \(\gamma_k\) from the SNR-balanced
schedule in Eq.~\ref{eq:snr-residual-weight}, content weight
\rev{\(0.15\) for LLFF and \(0.10\) for Tanks and Temples}, appearance learning
rate \(0.05\), and the fitting loss in Eq.~\ref{eq:fit-loss} with
\(\lambda_{\mathrm{ssim}}=0.2\), \(\lambda_{\mathrm{tv}}=0.01\), and
\(\lambda_{\mathrm{dino}}=0.01\). \rev{\polish{DReSG performs one latent Adam
update at each DDIM timestep, for a total of 10 updates per feedback stage;}} each feedback stage uses
30 inner steps for 3D residual fitting, and the final post color-transfer fitting
uses 50 additional steps. We apply color-gradient projection to \(\mathbf{c}\)
when fusing view-wise appearance updates. For performance analysis, we measure
optimization time, peak optimization memory, peak inference memory, and
inference FPS.

\subsection{Main Results}

\polish{Table~\ref{tab:main-quantitative} reports quality and efficiency metrics
for the final stylized 3D scene rather than only the selected active views. The
measurements support a multi-objective conclusion: DReSG attains the highest
DINO-C and the lowest short- and long-term drift, indicating strong content
preservation and persistent stylized appearance under viewpoint changes,
although its CLIP-S remains below the highest reported value. Its optimization
is faster than CLIPGaussian and SGSST and requires substantially less time than
FantasyStyle; at inference, it achieves the highest reported FPS and lowest
memory. These resource measurements make the intended trade-off explicit:
DReSG spends additional time constructing proposal-derived residual targets and
fitting them into the shared scene, but the final edited scene remains compact
and fast to render.}

\polish{Figure~\ref{fig:main-qualitative} compares how the methods reproduce the
style cues described in Section~\ref{sec:introduction}.} VGG-feature-based baselines often
preserve coarse layout but fragment directional strokes or coherent contours
into local texture responses. The diffusion-based baseline can generate stronger
style cues for region-wise color treatment or structured style details, but may
over-saturate colors, overwrite content, or detach details from object
boundaries. In these cases, DReSG better preserves object boundaries, occlusion
relationships, and foreground-background separation while retaining recognizable
style-reference cues. \polish{These observations align with the quantitative
results, which show strong content preservation and reduced view-dependent drift
alongside reference-style transfer.}

\subsection{Ablation Studies}
\label{sec:ablation}

We conduct component-wise ablations over four factors: attention-guided proposal
construction, SNR-based residual-strength modulation, shared-scene feedback, and
RGB-logit target construction. \polish{These ablations assess trade-offs rather
than identify a single-metric winner: some variants improve an isolated metric
by remaining closer to the input or weakening style, whereas the full model is
selected for its balance across style strength, content preservation, and view
stability.}
\rev{All comparisons hold the style image, selected active-view set, post
color-transfer configuration, and optimization budget constant.}
Figures~\ref{fig:residual-target-ablation},
\ref{fig:feedback-ablation}, and~\ref{fig:logit-ablation} provide qualitative
evidence for these factors, while Tables~\ref{tab:feedback-ablation},
\ref{tab:ablation}, and~\rev{\ref{tab:schedule-ablation}} summarize the
corresponding metrics.

\rev{For the schedule comparison, let \(F=N/S\) be the number of feedback
stages and \(k_n\) the scheduler state used at stage \(n\); the stage-wise scale
is \(\gamma_{k_n}=1+p_n\). SNR-balanced and Triangle use the scheduler
coordinate \(\bar{\alpha}_{k_n}\), with Triangle matching the range of
SNR-balanced but using a piecewise-linear profile. Timestep cosine is defined
over the feedback-stage index and places its peak at the midpoint of that index
range. The comparison also includes the fixed endpoints \(\gamma=1\) and
\(\gamma=2\).}

\begin{table}[t]
  \centering
  \definecolor{RankFirstBg}{RGB}{255,223,128} 
  \definecolor{RankSecondBg}{RGB}{220,224,229} 
  \providecommand{\rankfirst}[1]{\cellcolor{RankFirstBg}#1}
  \providecommand{\ranksecond}[1]{\cellcolor{RankSecondBg}#1}
  \caption{Attention-guided residual construction ablation. \(Q_c\) is the content query, and \(K_s,V_s\) are style keys and values. Metric abbreviations are defined in Section~\ref{sec:metrics}.}
  \label{tab:ablation}
  \resizebox{\linewidth}{!}{
    \begin{tabular}{lcccccc}
      \toprule
      Variant & C-S \(\uparrow\) & D-C \(\uparrow\) &
      ST-LP \(\downarrow\) & ST-RM \(\downarrow\) &
      LT-LP \(\downarrow\) & LT-RM \(\downarrow\) \\
      \midrule
      DReSG & \ranksecond{0.736} & 0.556 & \rankfirst{0.061} & \rankfirst{0.033} & \rankfirst{0.101} & \ranksecond{0.053} \\
      \midrule
      w/o \(Q_c\) & \rankfirst{0.779} & 0.407 & 0.063 & 0.035 & \ranksecond{0.104} & 0.057 \\
      w/o \(K_s\) & 0.655 & \ranksecond{0.625} & 0.064 & \rankfirst{0.033} & \ranksecond{0.104} & \ranksecond{0.053} \\
      w/o \(V_s\) & 0.594 & \rankfirst{0.695} & \ranksecond{0.062} & \rankfirst{0.033} & \rankfirst{0.101} & \rankfirst{0.050} \\
      \bottomrule
    \end{tabular}
  }
\end{table}

\begin{table}[t]
  \revcolor
  \centering
  \definecolor{RankFirstBg}{RGB}{255,223,128} 
  \definecolor{RankSecondBg}{RGB}{220,224,229} 
  \providecommand{\rankfirst}[1]{\cellcolor{RankFirstBg}#1}
  \providecommand{\ranksecond}[1]{\cellcolor{RankSecondBg}#1}
  \caption{\rev{Five-schedule residual-strength ablation under the same evaluation and optimization settings. Metric abbreviations are defined in Section~\ref{sec:metrics}.}}
  \label{tab:schedule-ablation}
  \resizebox{\linewidth}{!}{
    \begin{tabular}{lccc}
      \toprule
      Schedule & Extra scale \(p_n\) & C-S \(\uparrow\) & D-C \(\uparrow\) \\
      \midrule
      SNR-balanced & \(4\bar{\alpha}_{k_n}(1-\bar{\alpha}_{k_n})\) & \rankfirst{0.7362} & \rankfirst{0.5557} \\
      Fixed \(\gamma=1\) & \(0\) & 0.7291 & \ranksecond{0.5467} \\
      Fixed \(\gamma=2\) & \(1\) & 0.6617 & 0.5316 \\
      Triangle & \(1-|2\bar{\alpha}_{k_n}-1|\) & 0.7357 & 0.5370 \\
      Timestep cosine & \(\sin^2(\pi(n-1)/(F-1))\) & \ranksecond{0.7361} & 0.5277 \\
      \bottomrule
    \end{tabular}
  }
\end{table}

Table~\ref{tab:ablation} reports the attention-guided residual construction
ablation. Removing
\(K_s\) or \(V_s\) reduces CLIP-S, while removing \(Q_c\) increases CLIP-S but
substantially reduces DINO-C. The qualitative comparisons in
Fig.~\ref{fig:residual-target-ablation} illustrate these trade-offs,
supporting the complementary roles of the content query and style keys/values
rather than a single-metric ranking of the variants.

\rev{Table~\ref{tab:schedule-ablation} isolates residual-strength scheduling.
The three unimodal schedules yield nearly identical CLIP-S values. Among them,
SNR-balanced attains the
highest DINO-C and outperforms both fixed-scale baselines on both metrics. We
therefore adopt SNR-balanced; its smooth, scheduler-aligned profile is consistent
with the motivation in Section~\ref{sec:residual-targets} and achieves the best
measured style--content balance in this five-schedule comparison.}

\begin{table}[t]
  \centering
  \definecolor{RankFirstBg}{RGB}{255,223,128} 
  \definecolor{RankSecondBg}{RGB}{220,224,229} 
  \providecommand{\rankfirst}[1]{\cellcolor{RankFirstBg}#1}
  \providecommand{\ranksecond}[1]{\cellcolor{RankSecondBg}#1}
  \caption{Quantitative feedback ablation on the matched scene-style subset used for feedback variants.}
  \label{tab:feedback-ablation}
  \resizebox{\linewidth}{!}{
    \begin{tabular}{lcccccc}
      \toprule
      Variant & C-S \(\uparrow\) & D-C \(\uparrow\) &
      ST-LP \(\downarrow\) & ST-RM \(\downarrow\) &
      LT-LP \(\downarrow\) & LT-RM \(\downarrow\) \\
      \midrule
      DReSG & \rankfirst{0.7394} & \ranksecond{0.5592} &
      \ranksecond{0.0624} & \rankfirst{0.0339} &
      \ranksecond{0.1027} & \rankfirst{0.0548} \\
      w/o color-grad proj. & \ranksecond{0.7373} & 0.5549 &
      \rankfirst{0.0623} & \ranksecond{0.0341} &
      \rankfirst{0.1023} & \rankfirst{0.0548} \\
      w/o feedback & 0.6670 & \rankfirst{0.6773} &
      0.0718 & 0.0400 &
      0.1087 & 0.0611 \\
      \bottomrule
    \end{tabular}
  }
\end{table}

\begin{figure*}[!t]
  \centering
  \includegraphics[width=\textwidth]{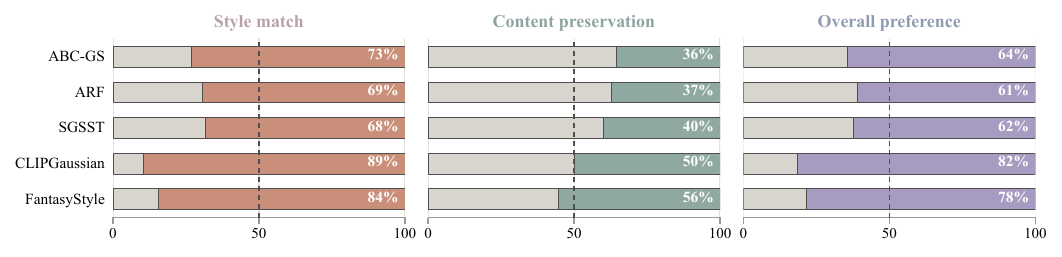}
  \caption{Pairwise user study. We report the percentage of votes selecting DReSG over each baseline for style match, content preservation, and overall preference.}
  \label{fig:user-study-chart}
\end{figure*}

\rev{The feedback ablation isolates the render-to-latent path that closes the
stage-wise loop. In the \emph{w/o feedback} variant, each active view retains its
current attention-optimized latent after 3D fitting instead of replacing it with
the VAE encoding of the updated render; residual-target construction and
shared-scene fitting are otherwise unchanged. Removing this path lowers style
strength and increases short- and long-term drift in
Table~\ref{tab:feedback-ablation}, while Fig.~\ref{fig:feedback-ablation} shows
local artifacts and view-inconsistent details.} \rev{Without color-gradient projection,
conflicting style updates from different views are fused directly. Although the
aggregate metrics remain close, the paired examples in
Fig.~\ref{fig:feedback-ablation} show greater local style fragmentation without
projection, qualitatively supporting its role in mitigating conflicting
appearance updates during multi-view fusion.}

\polish{The RGB-logit ablation compares the full model with a variant that scales
residuals directly in image space. Direct image-space extrapolation can produce
saturated or locally unstable targets; Fig.~\ref{fig:logit-ablation} shows that
removing RGB-logit scaling introduces localized color artifacts, whereas
RGB-logit construction yields cleaner, bounded appearance changes under the same
feedback setting. This comparison explains why the residual target is formed in
RGB-logit space before being fitted into the Gaussian scene.}

\subsection{User Study}

We conduct a pairwise user study to evaluate perceived style match, content
preservation, and overall quality. The study uses six scene-style pairs and
compares DReSG against five baselines with 30 participants. \polish{For each
baseline and criterion, this design yields 180 pairwise votes.} Trial order and
left-right order are randomized, and method names are hidden.

Participants answer three preference questions: which result has stronger style
expression, which result better preserves the original content, and which result
is preferred overall. This protocol separates the two main perceptual
objectives from the final trade-off, since a result can appear strongly stylized
while damaging the scene, or preserve content while appearing weakly stylized.
As shown in Fig.~\ref{fig:user-study-chart}, DReSG is preferred for style match
against all baselines, with rates from 68.33\% against SGSST to 89.44\% against
CLIPGaussian. \polish{Content-preservation votes reveal the style--content trade-off:}
conservative VGG-feature-based baselines can be favored for input preservation
alone, while DReSG reaches parity with CLIPGaussian and is preferred over
FantasyStyle. For overall preference, DReSG receives more than half of the votes
against every baseline, from 61.11\% against ARF to 81.67\% against
CLIPGaussian, suggesting that participants favor its combined style-transfer and
scene-preservation trade-off.

\section{Conclusion and Limitations}

This paper introduced DReSG for reference-guided stylization of scenes
represented by 3DGS.
DReSG converts attention-guided diffusion proposals into
proposal-render residual targets, constructs bounded RGB-logit fitting targets,
modulates residual strength with an SNR-balanced schedule, and fits the targets
to a shared Gaussian scene through multi-view rendering. \polish{Quantitative
results on LLFF and qualitative comparisons on LLFF and Tanks and Temples
\rev{show that} DReSG improves the balance between
reference-specific stylization, structure preservation, and cross-view
persistence compared with representative 3D stylization baselines; the ablation
results clarify how residual construction, residual modulation, and shared scene
fitting contribute to that balance.}

DReSG has three main limitations. First, \polish{its performance is limited by the proposal source}:
the quality and resolution of attention-guided diffusion proposals limit the
residual targets and thus the final stylized 3D scene. Stronger
high-resolution, multi-scale, or multi-view diffusion models may provide more
reliable proposal targets. Second, DReSG depends on
the quality of the input 3DGS. Reconstruction artifacts, missing geometry, or
weakly anchored regions can affect stylization stability, and
reconstruction-aware extensions could allocate Gaussians more effectively in
\polish{thin structures or regions containing fine details} or enrich their appearance and geometry
attributes. \polish{Extending DReSG to geometry reconstruction or repair remains
future work.} Sparse active views can also under-supervise rarely visible regions,
motivating adaptive view expansion for broader camera paths. Third, DReSG is
not feed-forward. Each scene-style pair still
requires iterative diffusion guidance and 3DGS optimization; distilling this
process into a scene/style-conditioned predictor for Gaussian attributes is a
promising direction.

\section*{Acknowledgments}

This work was supported by the National Natural Science Foundation of
China (Grant Nos. 62572194, 62472178, and 62376244) and the Shanghai Frontiers
Science Center of Molecule Intelligent Syntheses. This work was also supported
by the Key Technology Research and Development Program of the Shanghai Science
and Technology Commission (Grant Nos. 25511107200 and 25511102400).

\clearpage
\bibliographystyle{eg-alpha-doi} 
\bibliography{paper,egbibsample}       
\end{document}